\documentclass[11pt]{scaleai-paper}

\usepackage{amsmath}
\usepackage{amsfonts}
\usepackage{amssymb}
\usepackage{amsthm}
\usepackage{booktabs}
\usepackage{tabularx}
\usepackage{tabulary}
\usepackage{multirow}
\usepackage{subcaption}
\usepackage{float}
\usepackage{wrapfig}
\usepackage{nicefrac}
\usepackage{courier}
\usepackage[most]{tcolorbox}
\usepackage[square,numbers,sort&compress]{natbib}
\usepackage{xspace}
\usepackage{url}
\usepackage[colorlinks=true,linkcolor=scaleLink,citecolor=scaleLink,urlcolor=scaleLink]{hyperref}
\usepackage[capitalise,nameinlink]{cleveref}
\usepackage{fontawesome5}
\usepackage{upquote}
\usepackage{needspace}

\graphicspath{{./}{../}}

\allowdisplaybreaks

\newcommand{\bestcell}[1]{\cellcolor{scaleLavender!28}\textbf{#1}}

\newtcolorbox{promptbox}{
  breakable,
  enhanced,
  colback=scaleSage!12!white,
  colframe=scaleSage!70!black,
  boxrule=0.4pt,
  arc=2pt,
  left=6pt,
  right=6pt,
  top=6pt,
  bottom=6pt
}

\newtcolorbox{catbox}[1]{
  enhanced, breakable,
  colback=#1!10!white,
  colframe=#1!55!black,
  boxrule=0.45pt, arc=2pt,
  left=8pt, right=8pt, top=5pt, bottom=5pt,
  before skip=4pt, after skip=4pt,
}

\papertype{Scale AI Research}
\contact{\textcolor{scaleInk}{\faEnvelope}\ \texttt{utkarsh.tyagi@scale.com}}

\title{Not Every Rubric Teaches Equally:\\
Policy-Aware Rubric Rewards for RLVR}

\author[1]{Utkarsh Tyagi}
\author[1]{Xingang Guo}
\author[1]{MohammadHossein Rezaei}
\author[2]{Daniel George}
\author[1]{Anas Mahmoud}
\author[1]{Jackson Lee}
\author[1]{Bing Liu}
\author[1]{Yunzhong He}

\affil[1]{Scale AI}
\affil[2]{Persona}

\begin{document}

\maketitle

\begin{abstract}
Reinforcement learning with verifiable rewards has made post-training highly effective when correctness can be checked automatically. However, many important model behaviors require satisfying several qualitative criteria at once. Rubric-based rewards address this setting by grading prompt-specific criteria and aggregating them into a scalar reward. Yet standard static aggregations conflate a criterion's human-assigned importance with its current usefulness as an optimization signal. We show that this assumption breaks down in rubric RL: many important criteria are already saturated or currently unreachable, while criteria that distinguish rollouts are not necessarily those with the largest human weights. We introduce \textbf{POW3R}, a policy-aware rubric reward framework that preserves human weights and category balance as the rubric objective while adapting criterion-level reward weights during training. POW3R uses rollout-level contrast to emphasize criteria that currently separate the policy's outputs, making the GRPO reward more informative without changing the underlying evaluation target. Across three base policies on two datasets spanning multimodal and text-only settings, POW3R wins $24$ of $30$ base-policy/metric comparisons, improving both mean rubric reward and strict completion (the fraction of prompts whose response satisfies every \emph{required} rubric criterion) over vanilla GRPO with rubric rewards, and reaches the same plateau in $2.5$--$4\times$ fewer training steps. Rubric rewards should therefore distinguish what should matter in the final answer from what can teach the current policy.
\end{abstract}

\section{Introduction}
\label{sec:intro}

Reinforcement learning with verifiable rewards (RLVR) has become a central recipe for post-training language models on tasks where success can be cheaply and reliably checked. Group-relative methods such as GRPO have made this practical at scale by replacing a learned value model with within-prompt rollout comparison~\citep{deepseekmath, deepseek_r1, dapo}. The strength of the recipe is also its limitation: it works best when target behavior can be reduced to a single outcome score, and recent RLVR diagnostics and imperfect-verifier analyses already document that scalar rewards can hide heterogeneous failure modes and noise~\citep{spurious_rewards, noisy_verifiers}.

\begin{figure}[!t]
\centering
\includegraphics[width=\linewidth]{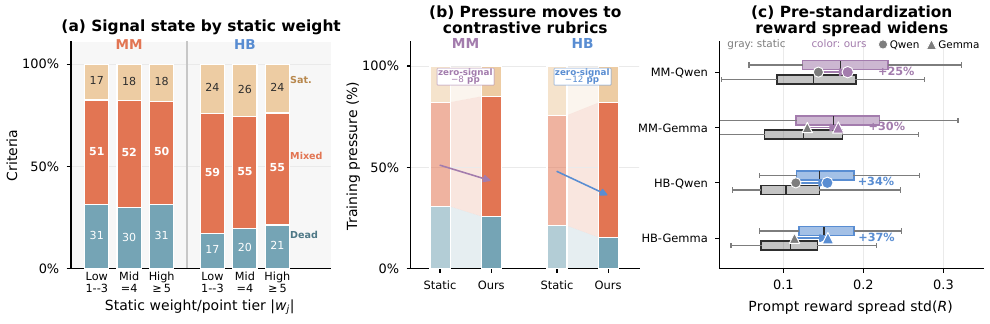}
\caption{\textbf{Rubric-pressure diagnostic.} We track each criterion's \emph{training pressure}\protect\footnotemark{} across criterion signal states. A criterion is \emph{dead} when no rollout passes it ($p_j{=}0$), \emph{saturated} when every rollout passes it ($p_j{=}1$), and \emph{mixed} when verdicts differ across the rollout group; dead and saturated criteria have $v_j{=}0$ and give no group-relative advantage signal. \textbf{(a)} Criteria grouped by setting and absolute static weight/point tier (Low: $|w_j|\!\in\!\{1,2,3\}$; Mid: $|w_j|\!=\!4$; High: $|w_j|\!\ge\!5$); high-weight criteria still carry substantial dead and saturated mass, so human importance is not the same as current learnability. \textbf{(b)} Within-category pressure before and after policy-aware reweighting, averaged over the two base policies per setting; pressure on zero-signal criteria drops by $8$--$12$ pp. \textbf{(c)} Prompt-level reward spread (rollout $\mathrm{std}(R)$ \emph{before} GRPO standardization), static (gray) vs policy-aware (colored). Mean spread widens by $25$--$37\%$, so fewer groups collapse to the tied $\mathrm{std}(R){=}0$ regime where every advantage is zero. MM/HB denote our multimodal dataset and HealthBench.}
\label{fig:diagnostic}
\end{figure}
\footnotetext{We define criterion $j$'s \emph{training pressure} as its within-category reward-weight share: $w_j / W_{\kappa_j}(q)$ under the static reward, $\tilde{w}_j^{(t)} / \tilde{W}_{\kappa_j}^{(t)}(q)$ under POW3R (\cref{eq:cat_reward,eq:rc_reward}).}

Many important behaviors do not collapse cleanly onto one outcome score. Long-form medical advice, scientific writing, coding help, and visually grounded reasoning are inherently multi-dimensional: a good answer must be factually correct, complete, faithful to evidence, well-formatted, and on-instruction at the same time. Expert rubric grading exposes this finer structure where exact-answer scoring is silent~\citep{healthbench}, and recent multimodal work documents that final-answer rewards can leave perception and grounding undertrained, with models sometimes ``reasoning past the image'' rather than from it~\citep{perception_r1, prco}. Open-ended quality is, in short, vector-valued; pushing post-training beyond strictly verifiable domains requires rewards that expose that vector rather than collapse it.

Rubrics provide that structure. A rubric decomposes response quality into prompt-specific criteria, each independently scored by an LLM judge, and rubric-based rewards have become a practical way to extend RL post-training beyond strictly verifiable domains~\citep{rubrics_as_rewards}, with text-only and multimodal rubric pipelines both growing rapidly~\citep{openrubrics, rubrichub, autorubric_r1v}. Rubrics, however, change the nature of reward design: it is no longer a verification problem but an \emph{aggregation} problem, since GRPO still requires a single scalar reward per rollout~\citep{deepseekmath} and every rubric criterion must eventually be folded into one number.

The common operational answer is a static weighted sum across rubric items~\citep{rubrics_as_rewards, healthbench}. This is convenient but contains a hidden assumption: that the human-assigned weight of a criterion expresses both its desired importance in the final answer \emph{and} its current usefulness as a training signal. The two are not the same. Under group-relative RL with outcome supervision, a criterion that every rollout passes, or that no rollout passes, adds the same constant to every reward and cancels out of the advantage; only criteria whose pass rate sits between the extremes can teach the current policy. A high-weight criterion can therefore be important for evaluation while still producing no gradient signal right now. This is the rubric-level form of a broader fixed-scalarization issue in multi-reward RL, where static weights preserve a target preference but route learning effort poorly across objectives~\citep{dynamic_reward_weighting}.

We test this assumption directly. Using two frozen base policies, Qwen3-VL-4B-Instruct~\citep{qwen3_vl} and Gemma~3 12B-IT~\citep{gemma3}, we sample rollout groups on $1{,}300$ prompts drawn from our multimodal dataset (\emph{MM}; \cref{sec:datasets}) and HealthBench English (\emph{HB})~\citep{healthbench}, judging every rubric criterion on every rollout with GPT-5.4-mini~\citep{openai_gpt54_mini_nano}.\footnote{The judge--effort combination is selected from a cost--quality calibration against a high-effort reference judge; the full agreement table is in \cref{app:judges}.} For each criterion $j$ we record its absolute weight $|w_j|$, pass rate $p_j$, variance $v_j{=}p_j(1{-}p_j)$, and training pressure.

The pattern is consistent across both policies and both settings (\cref{fig:diagnostic}). Roughly half of all rubric criteria are non-contrastive for a fresh policy: $17$--$26\%$ are saturated and $20$--$33\%$ are dead, leaving only the remaining half able to produce a contrastive gradient. The static aggregation therefore routes $45$--$51\%$ of within-category training pressure to criteria that cannot move the policy, and the problem is not confined to low-importance criteria: human weight and rollout variance are essentially uncorrelated, and roughly half of the highest-weight criteria already carry $v_j{=}0$. These shares change by only a few points across the four (model, dataset) combinations, so what we are seeing is a property of static aggregation, not of any single base policy or domain. \emph{Static weights tell us what should matter in the final answer, not which criteria can teach the current model.}

\begin{wrapfigure}[20]{r}{0.40\textwidth}
\centering
\vspace{-10pt}
\includegraphics[width=\linewidth]{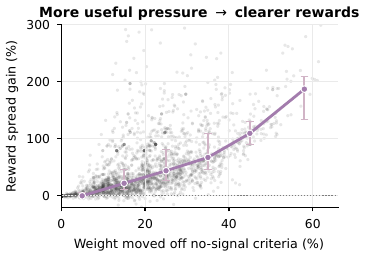}
\caption{\textbf{Mechanism check.} Each gray point is one prompt under one diagnostic run: Qwen3-VL-4B-Instruct or Gemma~3 12B-IT on our multimodal dataset (MM) or HealthBench English (HB). Moving weight off criteria that all rollouts pass or fail separates the rollout rewards more; reward spread is the standard deviation of rollouts before GRPO standardization. The trend holds prompt-by-prompt in every run.}
\label{fig:scatter}
\end{wrapfigure}

The diagnostic gives a direct design rule: preserve the evaluation rubric as the target, but route within-category pressure toward criteria that currently distinguish rollouts. This follows the multi-objective view that scalarization can be a training-time choice rather than only a fixed preference statement~\citep{roijers_mosdm}, and it complements multi-reward GRPO work showing that naive normalization can erase objective-specific signal~\citep{gdpo}. Our Policy-Aware Rubric Reward framework, \textbf{POW3R}, implements this rule on top of the standard rubric reward: (i) it measures each criterion's rollout contrastiveness from the smoothed standard deviation of its judge verdicts, (ii) blends and clips this signal into a bounded factor so saturated and uniformly failed criteria keep a learning floor while contrastive ones receive more pressure, and (iii) renormalizes within each rubric category so that the human weight prior and category mass remain intact. Offline replay confirms the local mechanism: POW3R moves pressure off dead and saturated criteria and widens the pre-standardization rollout reward spread (\cref{fig:diagnostic}b,c), and \cref{fig:scatter} verifies the same effect prompt by prompt.

\textbf{Key contributions.}
(i) We introduce a rubric-pressure diagnostic that exposes how static rubric aggregation routes training pressure, and use it to show that human-assigned importance and current policy learnability decouple in rubric RL.
(ii) We propose \textbf{POW3R}, a policy-aware rubric reward that preserves human weights and category balance while reallocating within-category training pressure to currently informative criteria.
(iii) Under the GRPO recipe across three base policies on each of MM and HealthBench, POW3R beats binary, static-scalar, and category-balanced rewards on $24$ of $30$ comparisons, matches them in $2.5$--$4\times$ fewer steps, and preserves external VLM benchmark scores.

\section{Related work}
\label{sec:related}

\paragraph{Rubric-based rewards and the policy-aware view.}
Rubric-based rewards extend RL post-training beyond deterministic verifiers by decomposing response quality into prompt-specific criteria scored by an LLM judge \citep{rubrics_as_rewards}. Expert-written rubric benchmarks scale this signal in medicine \citep{healthbench} and have begun to reach other modalities such as multi-turn spoken dialogue \citep{audio_multichallenge}, while synthetic or semi-automatic pipelines reduce rubric-authoring cost \citep{openrubrics, rubrichub}. Other work modifies the rubric set during training: \citet{online_rubrics} elicit rubrics from pairwise comparisons, \citet{dr_tulu} co-evolve rubrics with the policy for long-horizon generation, and \citet{autorubric_r1v} generate multimodal rubric rewards from successful trajectories. Closest to ours is \citet{rucl}, which stratifies generalized rubrics into a perception-to-reasoning curriculum and dynamically reweights them across training; we share the diagnosis that not all criteria are equally learnable at every stage, but our rubrics are prompt-specific and human-authored, we preserve human-assigned importance via static within-category weights, and we derive dynamic factors per prompt from the current policy's rollout variance rather than from a global capability schedule.

\vspace{-8pt}

\paragraph{Multi-reward RL and multimodal RLVR.}
Several lines treat alignment as multi-objective rather than scalar optimization \citep{morlaif, dynamic_reward_weighting, roijers_mosdm}, complementing RLHF/RLAIF recipes that compress rich human or AI feedback into a single reward target whose scalar form can hide heterogeneous values and failure modes \citep{ziegler_human_preferences, stiennon_summarize, instructgpt, constitutional_ai, rlaif, rewardbench}. \citet{gdpo} show that naively normalizing multi-reward rollouts under GRPO collapses distinct reward combinations into identical advantages. On the multimodal side, RLVR extends GRPO-style post-training to vision-language reasoning \citep{visual_rft, vision_r1, r1_vl, ground_r1} and adds visual perception rewards, evidence gates, dense spatial rewards, or token-level reweighting when final-answer signals underfit perception \citep{perception_r1, pearl, rewardmap, token_reweighting}, while complementary benchmarks evaluate tool-enabled image perception, transformation, and reasoning under a unified protocol \citep{visualtoolbench}. RLVR diagnostics in parallel show that observed gains can reflect spurious signals rather than newly learned capabilities, motivating inspection at the criterion level \citep{spurious_rewards, rlvr_reasoning_capacity}.

\section{Preliminaries}
\label{sec:background}

\subsection{Group relative policy optimization}
\label{sec:grpo}

We post-train policies with Group Relative Policy Optimization (GRPO)~\citep{deepseekmath}, the algorithm underlying recent reasoning-RL recipes \citep{deepseek_r1, dapo}. For each prompt $q \sim P(Q)$, GRPO samples a group of $G$ outputs $\{o_1, \ldots, o_G\}$ from the old policy $\pi_{\theta_{\text{old}}}$ and optimizes the policy by maximizing
\begin{multline}
\mathcal{J}_{\text{GRPO}}(\theta) \;=\; \mathbb{E}_{q,\,\{o_i\}\sim \pi_{\theta_{\text{old}}}(\cdot\mid q)}\!\Bigg[\frac{1}{G}\sum_{i=1}^{G}\frac{1}{|o_i|}\sum_{t=1}^{|o_i|}\Big\{\min\!\big[r_{i,t}\hat{A}_{i,t},\\
\mathrm{clip}(r_{i,t},\, 1{-}\varepsilon,\, 1{+}\varepsilon)\,\hat{A}_{i,t}\big] \;-\; \beta\, \mathbb{D}_{\text{KL}}\big[\pi_\theta\,\|\,\pi_{\text{ref}}\big]\Big\}\Bigg],
\label{eq:grpo}
\end{multline}
where $r_{i,t}(\theta) = \pi_\theta(o_{i,t}\mid q, o_{i,<t}) / \pi_{\theta_{\text{old}}}(o_{i,t}\mid q, o_{i,<t})$ is the per-token probability ratio. Writing $u_{i,t} = \pi_{\text{ref}}(o_{i,t}\mid q, o_{i,<t}) / \pi_\theta(o_{i,t}\mid q, o_{i,<t})$ for the reference-to-policy ratio, the per-token Schulman~k3 estimator is
\begin{equation}
\mathbb{D}_{\text{KL}}\!\big[\pi_\theta\,\big\|\,\pi_{\text{ref}}\big] \;=\; u_{i,t} \;-\; \log u_{i,t} \;-\; 1.
\label{eq:kl}
\end{equation}
We use outcome supervision: a scalar reward $R(o_i; q)$ is assigned to each output $o_i$, standardized within the group, and the resulting $\hat{A}_{i,t} = (R(o_i; q) - \mathrm{mean}(\mathbf{R}))/\mathrm{std}(\mathbf{R})$ with $\mathbf{R} = \{R(o_j;q)\}_{j=1}^{G}$ is broadcast to every token in $o_i$. When $\mathrm{std}(\mathbf{R}){=}0$ (all rollouts tied) we set $\hat{A}_{i,t}{=}0$, so the group contributes no gradient that step -- a regime the \cref{sec:intro} diagnostic shows is reached on a non-trivial share of prompts under static aggregation. The construction of $R(o_i; q)$ is the focus of this paper.\label{eq:advantage}

\subsection{Rubric-based rewards}
\label{sec:rubrics}

A rubric-based reward decomposes response quality into prompt-specific criteria scored by an LLM judge \citep{rubrics_as_rewards, healthbench}. For a prompt $q$ we write its rubric set as $\mathcal{R}(q) = \{(c_j, w_j, \kappa_j)\}_{j=1}^{N(q)}$, where each criterion $c_j$ has a static human weight $w_j \in \mathbb{N}_{>0}$ and a category label $\kappa_j \in \{1, \ldots, K\}$; let $\mathcal{C}_k(q) = \{j : \kappa_j = k\}$. The grader produces $s_j(o, q) \in [0, 1]$, with $s_j(o, q) = 1$ meaning response $o$ satisfies $c_j$. The standard rubric reward used in prior work is the static weighted sum $R_{\text{scalar}}(o; q) = \sum_{j=1}^{N(q)} w_j\, s_j(o, q)$. This sum bakes in three implicit assumptions: (i) categories contain comparable numbers of criteria; (ii) criteria within a category are similarly informative under the current policy; and (iii) each $w_j$ expresses \emph{both} end-state importance and current training usefulness. The next section relaxes (i)--(iii) while leaving $s_j(\cdot,\cdot)$ and the human weights $w_j$ unchanged.

\section{Method}
\label{sec:method}
POW3R changes only the reward aggregation before GRPO standardization, keeping the rubric, judge scores, and human weights fixed while reallocating within-category pressure toward criteria that distinguish the current rollout group.

\vspace{-10pt}

\paragraph{Category-normalized baseline.}
For each prompt, let $W_k(q)=\sum_{j\in\mathcal{C}_k(q)} w_j$ and $K_q$ be the number of populated categories. Define
\begin{equation}
R_{\text{cat}}(o; q) = \frac{1}{K_q}\sum_{k:\, \mathcal{C}_k(q) \neq \emptyset}\frac{1}{W_k(q)}\sum_{j \in \mathcal{C}_k(q)} w_j\, s_j(o, q).
\label{eq:cat_reward}
\end{equation}

\vspace{-10pt}

\paragraph{Policy-aware factors.}
Each prompt-rubric factor $\alpha_j^{(t)}$ starts at $1$ and is applied to all $G$ rollouts in epoch $t$; after the epoch, judge calls yield each criterion's pass rate and variance,
\begin{equation}
p_j^{(t)} \;=\; \frac{1}{n_j}\sum_{i\in\mathcal{V}_j} s_j(o_i, q), \qquad
v_j^{(t)} \;=\; \frac{1}{n_j}\sum_{i\in\mathcal{V}_j}\!\big(s_j(o_i, q) - p_j^{(t)}\big)^2.
\label{eq:variance}
\end{equation}
with $\mathcal{V}_j$ the valid-verdict set and $n_j{=}|\mathcal{V}_j|$ (criteria with $<\!\lceil0.75G\rceil$ valid verdicts retain their previous factor). POW3R then smooths the variance, category-normalizes it, blends toward $1$, clips, and EMA-updates:
\begin{align}
g_j^{(t)} \;&=\; \sqrt{v_j^{(t)}+\epsilon}, \qquad
\bar g_k^{(t)} \;=\; \frac{\sum_{j'\in\mathcal{C}^{\mathrm{valid}}_k(q)} w_{j'}\,g_{j'}^{(t)}}{\sum_{j'\in\mathcal{C}^{\mathrm{valid}}_k(q)} w_{j'}}, \label{eq:signal}\\[2pt]
\rho_j^{(t)} \;&=\; \frac{g_j^{(t)}}{\bar g_{\kappa_j}^{(t)}}, \qquad
\hat{\alpha}_j^{(t)} \;=\; \operatorname{clip}\!\left((1-\lambda)+\lambda\rho_j^{(t)},\, \alpha_{\min},\, \alpha_{\max}\right), \label{eq:alpha}\\[2pt]
\alpha_j^{(t+1)} \;&=\; \operatorname{clip}\!\left((1-\beta_{\mathrm{ema}})\alpha_j^{(t)}+\beta_{\mathrm{ema}}\hat{\alpha}_j^{(t)},\, \alpha_{\min},\, \alpha_{\max}\right).
\label{eq:ema}
\end{align}
If all valid signals in a category vanish, POW3R sets $\hat{\alpha}_j^{(t)}{=}1$; $\lambda$ trades off prior vs.\ rollout contrast, $\beta_{\mathrm{ema}}$ sets response speed, $[\alpha_{\min}, \alpha_{\max}]$ bound deviation from $1$, and $\epsilon{>}0$ stabilizes ratios.

\vspace{-10pt}

\paragraph{POW3R reward.}
At epoch $t$, set $\tilde{w}_j^{(t)}=w_j\alpha_j^{(t)}$ and $\tilde{W}_k^{(t)}(q)=\sum_{j\in\mathcal{C}_k(q)}\tilde{w}_j^{(t)}$, then compute
\begin{equation}
R_{\text{dyn}}^{(t)}(o; q) = \frac{1}{K_q}\sum_{k:\, \mathcal{C}_k(q) \neq \emptyset}\frac{1}{\tilde{W}_k^{(t)}(q)}\sum_{j \in \mathcal{C}_k(q)} \tilde{w}_j^{(t)}\, s_j(o, q).
\label{eq:rc_reward}
\end{equation}
\Cref{eq:rc_reward} keeps category mass uniform and uses $w_j$ as prior: if factors in a category are equal, we get \cref{eq:cat_reward}. The same $\{\alpha_j^{(t)}\}$ are used for all $G$ rollouts and fed into GRPO, requiring no optimizer change.

\section{Experimental setup}
\label{sec:setup}

\begin{wraptable}{r}{0.39\linewidth}
\vspace{-12pt}
\centering
\caption{Multimodal rubric-RL dataset statistics. The six rubric categories are the high-level quality dimensions used by the reward.}
\vspace{6pt}
\label{tab:dataset}
\footnotesize
\setlength{\tabcolsep}{3pt}
\renewcommand{\arraystretch}{1.07}
\begin{tabular*}{\linewidth}{@{\extracolsep{\fill}}lr@{}}
\toprule
\multicolumn{2}{l}{\textbf{Tasks}} \\
Total / train / dev / test                       & 10K / 8K / 1K / 1K \\
Single / Multi turn                              & 65\% / 35\% \\
\midrule
\multicolumn{2}{l}{\textbf{Rubrics}} \\
Total criteria                                   & 84{,}403 \\
Mean / median per task                           & 8.4 / 7 \\
Human weight $w_j$                               & integer in $\{1,\ldots,5\}$ \\
\midrule
\multicolumn{2}{l}{\textbf{Rubric category} (share, $\bar w$)} \\
Visual perception              & 32.4\%,\, 3.40 \\
Visual reasoning               & 26.0\%,\, 3.74 \\
Content completeness           & 16.9\%,\, 3.10 \\
Instruction following          & 11.4\%,\, 3.77 \\
Truthfulness                   & 10.5\%,\, 3.58 \\
Style / presentation           & \phantom{0}2.8\%,\, 2.30 \\
\bottomrule
\end{tabular*}
\vspace{-10pt}
\end{wraptable}

\begin{figure}[!t]
\centering
\includegraphics[width=0.9\linewidth]{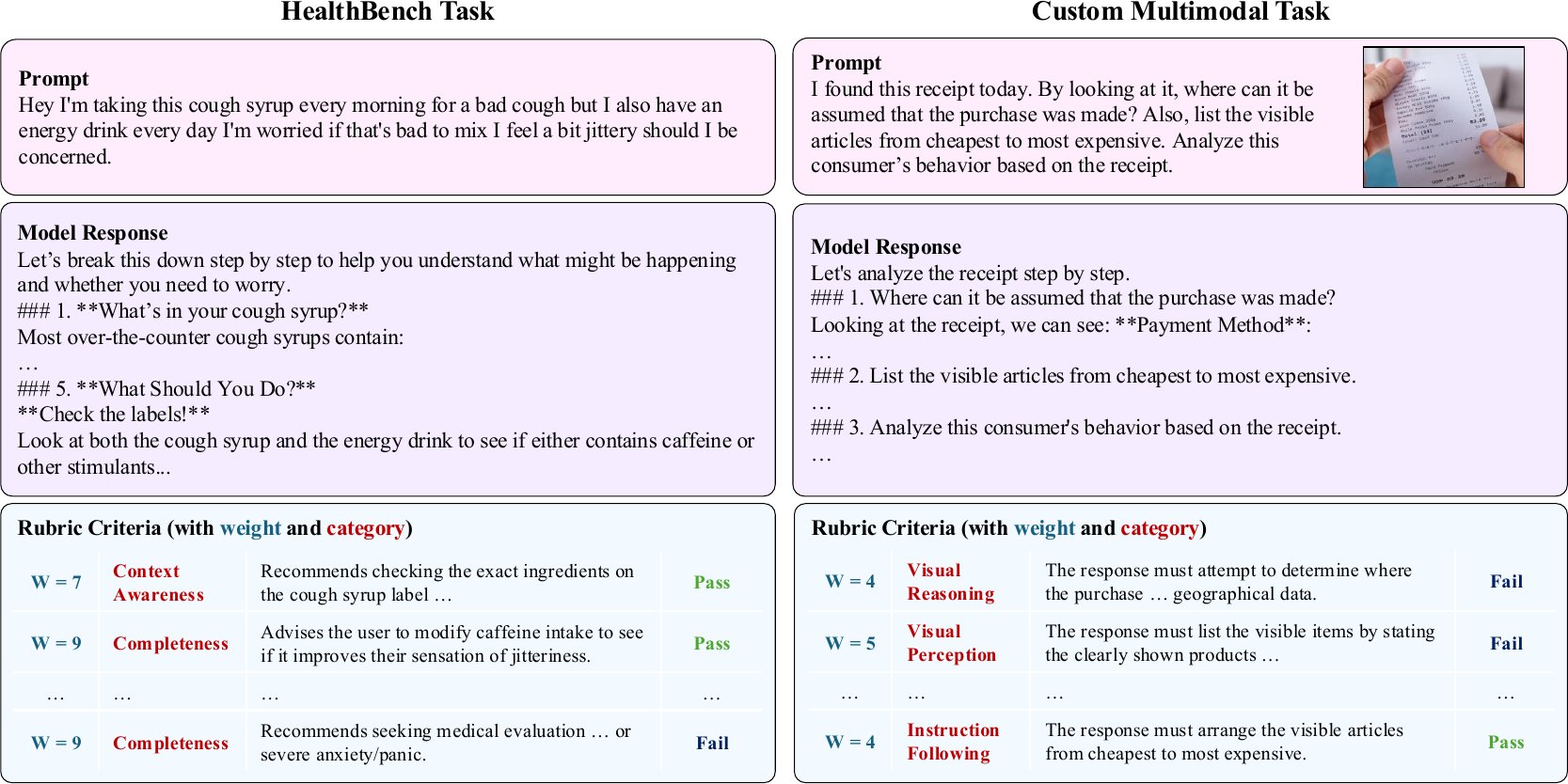}
\caption{\textbf{Illustrative tasks from each rubric-RL setting.} Each task carries a prompt, a sampled response, and a prompt-specific rubric checklist; MM includes image input, while HealthBench is text-only. Reward judging uses this same criterion-level interface: each listed criterion receives an independent binary verdict before aggregation.}
\vspace{-8pt}
\label{fig:examples}
\end{figure}

\paragraph{Datasets.}\label{sec:datasets}
We choose datasets that expose criterion-level categories and static importance weights rather than only a single outcome label \citep{rubrics_as_rewards, healthbench}. \textbf{HB} is HealthBench~\citep{healthbench} restricted to English-language prompts, with native physician-authored point-valued criteria; we use HealthBench's $500$-task \emph{hard} subset as the test split and a separate $10\%$ slice of the remaining English training prompts as the dev split. More details on HealthBench in \cref{app:hb_signed}. \textbf{MM} is our $10$k-task multimodal dataset, selected from a contributor-authored prompt pool because existing rubric-RL datasets do not simultaneously provide complex images, prompt-specific categories, static weights, and enough scale for generalisability. Each MM task pairs an image with a prompt and a rubric set spanning six quality categories (\cref{tab:dataset}); the images span charts, diagrams, photos, screenshots, and natural scenes, and each rubric criterion is anchored to specific visual elements or prompt instruction during authoring. \cref{fig:examples} illustrates the shared rubric-RL setting, and \cref{app:rubric_annotations} gives annotation details.

\label{app:examples}

\vspace{-10pt}

\paragraph{Models.}
On MM we post-train three vision-language base policies: Qwen3-VL-4B-Instruct and Qwen3-VL-8B-Instruct~\citep{qwen3_vl}\footnote{Qwen release pages: \href{https://hf.co/collections/Qwen/qwen3-vl}{Qwen3-VL}; \href{https://hf.co/collections/Qwen/qwen3}{Qwen3}.}, and Gemma~3 4B-IT~\citep{gemma3}\footnote{Gemma~3 release page: \href{https://hf.co/collections/google/gemma-3-release}{Gemma~3}.}. On HB we post-train three text-only base policies: Qwen3-4B-Instruct-2507 and Qwen3-8B~\citep{qwen3}, and Gemma~3 4B-IT~\citep{gemma3}. The diagnostic of \cref{sec:intro} additionally uses Qwen3-VL-4B-Instruct and the larger Gemma~3 12B-IT to check that the findings are not specific to a single base model.

\vspace{-10pt}

\paragraph{Reward judging.}
The reward judge is queried \emph{per rubric criterion}: every (prompt, rollout, criterion) triple gets a reasoning-then-verdict call, returning a one-sentence rationale and a binary $\{0,1\}$ judgment for aggregation. Training rewards use GPT-5.4-nano with medium-effort reasoning and explanations; held-out evaluation responses are re-scored by GPT-5.4-mini with the same reasoning setting to reduce judge--training entanglement~\citep{openai_gpt54_mini_nano}. \Cref{app:judges} gives the cost--agreement calibration and shows why verdict-only and per-category batched judges were not used. Both judges run at temperature $T{=}1.0$ with up to $2048$ completion tokens; the system prompt is reproduced in \cref{app:prompts}.

\vspace{-10pt}

\paragraph{Baselines.}
We use \textbf{POW3R} for the framework and $R_{\text{dyn}}$ for the scalar reward it sends to GRPO. We compare five post-training settings, all using the same rubric set and judge. \textit{(i)} \textbf{Base model}: the un-trained checkpoint, used as the no-RL reference. \textit{(ii)} \textbf{Binary}: a sparse all-or-nothing reward, $R_{\text{binary}}(o;q){=}\mathbf{1}\!\left[\text{every required criterion passes}\right]$ on MM and $R_{\text{binary}}(o;q){=}\mathbf{1}\!\left[\text{HealthBench scorer}{=}1\right]$ on HB; included as the exact-answer-style RLVR baseline. \textit{(iii)} \textbf{Static scalar}: the standard prior-work weighted sum $R_{\text{scalar}}$ from \cref{sec:rubrics}. \textit{(iv)} \textbf{Category-balanced}: the static category-balanced reward $R_{\text{cat}}$ from \cref{eq:cat_reward}. \textit{(v)} \textbf{POW3R dynamic}: $R_{\text{dyn}}$ from \cref{eq:rc_reward}. Each reported trained setting averages three completed runs under the same split, decoding, and evaluation protocol. The trained settings all run the same GRPO objective (\cref{sec:grpo}); only the rubric aggregation changes between them.

\vspace{-10pt}

\paragraph{Evaluation.}
At evaluation time, each completed policy is decoded on held-out prompts and every response is re-scored by the held-out judge. We report mean rubric reward on the $1{,}000$-task MM test set and HealthBench's $500$-task hard test split. For MM the rubric reward is the static weighted aggregation $R_{\text{scalar}}$ from \cref{sec:rubrics} normalized to $0$--$100\%$, applied uniformly across all five reward constructions so the evaluation target is held fixed; for HB we use the HealthBench's official scoring script. We also report \emph{strict completion} -- the fraction of prompts whose response satisfies every criterion flagged as \emph{required} in the rubric, and per-category mean pass rate. Rubric reward measures average quality under the rubric; strict completion measures all-required-criterion success with no partial credit.

\textbf{Transfer benchmarks (MM only).} To check that POW3R does not over-fit the rubric judge, we also evaluate the trained MM policies on six external VLM benchmarks: HallusionBench~\citep{hallusionbench}, POPE~\citep{pope}, MM-IFE~\citep{mmifengine}, MMVetV2~\citep{mmvet_v2}, MathVista~\citep{mathvista}, and RealWorldQA~\citep{realworldqa}. 

\vspace{-10pt}

\paragraph{Configuration.}
GRPO runs with $G{=}16$ rollouts per prompt-group, sampling temperature $T{=}1.0$, and a maximum completion length of $3584$ tokens. We use a learning rate of $3\!\times\!10^{-7}$, KL coefficient $\beta{=}0.1$, clip range $\varepsilon{=}0.2$, and $\mathrm{max\_grad\_norm}{=}0.5$, with a per-device batch size of $1$ and $4$ gradient-accumulation steps under DeepSpeed ZeRO-3~\citep{deepspeed_zero} in BF16 with gradient checkpointing. All training runs use one node with $8{\times}$H100 GPUs and run for up to $664$ GRPO steps. The dynamic-factor parametrization (Eqs.~(\ref{eq:signal})--(\ref{eq:ema})) uses $\alpha_{\min}{=}0.67$, $\alpha_{\max}{=}1.5$, $\epsilon{=}10^{-4}$, smoothing weight $\lambda{=}0.5$, EMA coefficient $\beta_{\mathrm{ema}}{=}0.2$, and minimum valid rollout fraction $0.75$ for the completed POW3R dynamic run.

\section{Results and analysis}
\label{sec:results}

\subsection{Main results}
\label{sec:main_results}

\begin{table}[!t]
\centering
\caption{\textbf{Held-out evaluation on MM test and selected external VLM benchmarks.} Entries are percentages: rubric reward is the mean held-out rubric score with partial credit, while strict completion is the fraction of prompts whose response satisfies every required criterion. External columns report HallusionBench, POPE, MM-IFE, MMVetV2, MathVista, and RealWorldQA. Best entries within each base-policy block are in bold and shaded.}
\label{tab:mm_main_results}
\scriptsize
\renewcommand{\arraystretch}{1.05}
\setlength{\tabcolsep}{1.7pt}
\begin{tabular*}{\textwidth}{@{\extracolsep{\fill}}llcccccccc@{}}
\toprule
 & & \multicolumn{2}{c}{\textbf{MM test}} & \multicolumn{6}{c}{\textbf{VLM benchmarks}} \\
\cmidrule(lr){3-4}\cmidrule(lr){5-10}
\textbf{Base policy} & \textbf{Setting} & \textbf{Rubric} & \textbf{Strict comp.} &
\textbf{HallusionBench} & \textbf{POPE} & \textbf{MM-IFE} & \textbf{MMVetV2} & \textbf{MathVista} & \textbf{RealWorldQA} \\
\midrule
\multirow{5}{*}{Qwen3-VL-4B} & Base & 42.6 & 14.7 & 62.1 & 88.8 & 11.5 & 49.9 & 71.0 & 71.9 \\
 & Binary & 42.8 & 14.6 & 62.0 & 88.6 & 11.8 & 49.8 & 71.2 & 71.7 \\
 & Static scalar & 47.1 & 17.9 & 62.2 & \bestcell{89.2} & 12.5 & 51.0 & 72.2 & 72.4 \\
 & Category-balanced & 47.9 & 18.7 & \bestcell{62.5} & 89.0 & 13.0 & 51.8 & 72.8 & 72.8 \\
 & \textbf{POW3R dynamic} & \bestcell{48.8} & \bestcell{20.2} & 62.4 & 89.1 & \bestcell{13.5} & \bestcell{52.2} & \bestcell{73.2} & \bestcell{73.1} \\
\addlinespace[2pt]
\multirow{5}{*}{Qwen3-VL-8B} & Base & 46.8 & 18.4 & 62.4 & 88.4 & 12.5 & 49.7 & 73.6 & 68.6 \\
 & Binary & 47.0 & 18.3 & 62.5 & 88.2 & 12.3 & 49.5 & 73.8 & 69.0 \\
 & Static scalar & 49.5 & 20.7 & 62.2 & 88.6 & 13.5 & 51.0 & 74.1 & 69.4 \\
 & Category-balanced & 50.4 & 21.6 & 62.6 & 88.7 & 14.2 & 51.8 & \bestcell{74.8} & 69.7 \\
 & \textbf{POW3R dynamic} & \bestcell{51.6} & \bestcell{22.6} & \bestcell{62.9} & \bestcell{88.9} & \bestcell{14.8} & \bestcell{52.6} & 74.6 & \bestcell{70.2} \\
\addlinespace[2pt]
\multirow{5}{*}{Gemma3-4B} & Base & 34.1 & 7.7 & 53.9 & 82.6 & 6.0 & 38.9 & 47.5 & 51.0 \\
 & Binary & 33.9 & 7.6 & 54.0 & 82.5 & 6.1 & 38.8 & 47.6 & 50.8 \\
 & Static scalar & 35.8 & 8.5 & 53.4 & 82.7 & 6.3 & 39.5 & 48.0 & 51.4 \\
 & Category-balanced & 36.7 & 9.4 & 54.1 & 82.8 & 6.7 & 40.2 & \bestcell{49.8} & 51.8 \\
 & \textbf{POW3R dynamic} & \bestcell{37.8} & \bestcell{10.3} & \bestcell{54.4} & \bestcell{83.0} & \bestcell{7.2} & \bestcell{40.8} & 49.6 & \bestcell{52.3} \\
\bottomrule
\end{tabular*}
\end{table}

\begin{wraptable}{r}{0.45\textwidth}
\vspace{-12pt}
\centering
\caption{\textbf{HealthBench English test split.} Entries are percentages except $\Delta$, which is already in pp. Best cells within each base-policy block are shaded.}
\label{tab:hb_results}
\scriptsize
\setlength{\tabcolsep}{2.0pt}
\renewcommand{\arraystretch}{1.0}
\resizebox{\linewidth}{!}{%
\begin{tabular}{llccc}
\toprule
\textbf{Base} & \textbf{Setting} & \textbf{Overall} & \textbf{$\Delta$} & \textbf{Strict} \\
\midrule
\multirow{5}{*}{Qwen3-4B} & Base & 28.0 & -- & 4.2 \\
 & Binary & 27.8 & $-0.2$ & 4.0 \\
 & Static scalar & 29.6 & $+1.6$ & 3.8 \\
 & Category-balanced & 30.5 & $+2.5$ & 4.6 \\
 & \textbf{POW3R dynamic} & \bestcell{32.7} & \bestcell{$+4.7$} & \bestcell{5.4} \\
\addlinespace[1pt]
\multirow{5}{*}{Qwen3-8B} & Base & 27.8 & -- & 4.4 \\
 & Binary & 27.9 & $+0.1$ & 4.4 \\
 & Static scalar & 29.9 & $+2.1$ & 4.5 \\
 & Category-balanced & 30.7 & $+2.9$ & \bestcell{4.8} \\
 & \textbf{POW3R dynamic} & \bestcell{31.5} & \bestcell{$+3.7$} & 4.6 \\
\addlinespace[1pt]
\multirow{5}{*}{Gemma3-4B} & Base & 24.3 & -- & 2.4 \\
 & Binary & 24.2 & $-0.1$ & 2.2 \\
 & Static scalar & 25.4 & $+1.1$ & \bestcell{2.8} \\
 & Category-balanced & 27.1 & $+2.8$ & 2.6 \\
 & \textbf{POW3R dynamic} & \bestcell{28.1} & \bestcell{$+3.8$} & 2.7 \\
\bottomrule
\end{tabular}
}%
\vspace{-25pt}
\end{wraptable}

\Cref{tab:mm_main_results,tab:hb_results} compare POW3R with the base model, binary reward, static scalar reward, and category-balanced reward under the same GRPO setup. Our key findings are:
\begin{enumerate}[leftmargin=*, itemsep=2pt, topsep=2pt]
\item \textbf{POW3R is the strongest reward on the main rubric objectives across both datasets.} Across the two main-results tables, POW3R achieves the best score on $24$ of the $30$ base-policy/metric comparisons, sweeping every MM rubric-reward and strict-completion column in \cref{tab:mm_main_results} and every HealthBench overall-reward column in \cref{tab:hb_results}. The six non-POW3R cells are split between external VLM benchmarks (which we do not train against directly; 4 of 6) and the HB strict perfect-score column, where POW3R is best on Qwen3-4B but trails the static or category-balanced reward by $0.1$--$0.2$ pp on Qwen3-8B and Gemma3-4B.
\end{enumerate}
\WFclear
\begin{enumerate}[leftmargin=*, itemsep=2pt, topsep=0pt, start=2]
\item \textbf{Per-rubric-category analysis on MM: POW3R's gain is consistent across categories, with the largest jumps on the contrastive ones.} A separate per-category analysis (see \cref{fig:learning_dynamics} for the full trajectories) shows that on Qwen3-VL-4B, POW3R leads on every rubric category for the full training schedule. The biggest gaps over the static baselines appear on Visual Perception, Visual Reasoning, Truthfulness, Content, and Instruction Following; on Writing Style the three rewards stay within roughly a point of each other because most Writing Style criteria are already passed by the base policy. This is by design: POW3R concentrates pressure where the rollout group exposes learnable disagreement, and reduces to the static baseline on categories with no remaining contrast to exploit.
\end{enumerate}

\vspace{-4pt}
\begin{figure}[!t]
\centering
\includegraphics[width=0.95\linewidth]{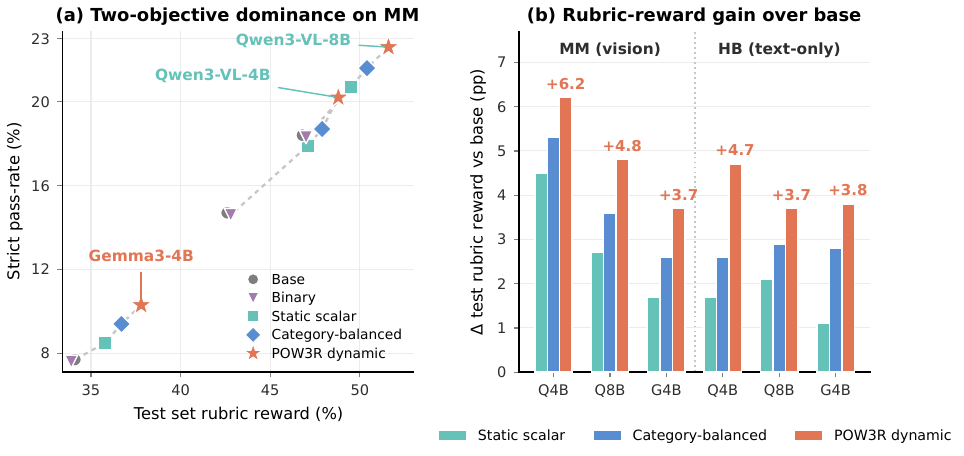}
\vspace{-8pt}
\caption{\textbf{Main result visual summary.} \textbf{(a)} MM test rubric reward and strict completion for each reward construction; lines connect methods trained from the same base policy. \textbf{(b)} Test rubric-reward gain over the corresponding base model on MM and HB.}
\vspace{-10pt}
\label{fig:main_visual_summary}
\end{figure}

\begin{enumerate}[leftmargin=*, itemsep=2pt, topsep=0pt, start=3]
\item \textbf{Two-objective dominance view separates mean quality from all-criteria success.} \Cref{fig:main_visual_summary}a places each method in the MM test rubric-reward/strict-completion plane; POW3R is the top-right endpoint of every base-policy line, so it Pareto-dominates the other four constructions on both objectives. This matters because a higher mean rubric score can still leave prompts with one failed required criterion; the strict-completion axis tests whether partial-credit gains turn into complete rubric satisfaction.
\item \textbf{$\Delta$ vs base across all six base policies shows cross-setting consistency.} \Cref{fig:main_visual_summary}b summarizes test gains in both modalities: $R_{\text{dyn}} > R_{\text{cat}} > R_{\text{scalar}} > 0$ in all six setting/base-policy combinations, and the smallest POW3R gain is still $+3.7$ pp. The ordering is unchanged between the multimodal and text-only settings, suggesting that POW3R is not exploiting one dataset's rubric convention.
\end{enumerate}

\subsection{Training efficiency: steps to a target reward}
\label{sec:efficiency}

\begin{wraptable}{r}{0.49\textwidth}
\vspace{-12pt}
\centering
\caption{\textbf{First step crossing each validation reward threshold (Qwen3-VL-4B, MM).} Thresholds are percentages; ``--'' means not reached within $664$ steps. Validation reward is the same rubric metric reported on the test split in \cref{tab:mm_main_results}, but evaluated on the dev split. }
\vspace{3pt}
\label{tab:convergence}
\scriptsize
\setlength{\tabcolsep}{3pt}
\renewcommand{\arraystretch}{0.98}
\resizebox{\linewidth}{!}{%
\begin{tabular}{ccccc}
\toprule
\textbf{Threshold} & \textbf{Static} & \textbf{Category-balanced} & \textbf{POW3R} & \textbf{Speed-up} \\
\midrule
$45.0$ & $249$ & $166$ & \bestcell{83}  & $3.0\times$ \\
$46.0$ & $249$ & $332$ & \bestcell{83}  & $4.0\times$ \\
$47.0$ & $415$ & $415$ & \bestcell{166} & $2.5\times$ \\
$48.0$ & $581$ & $581$  & \bestcell{166} & $3.5\times$ \\
$49.0$ & --    & --    & \bestcell{249} & --          \\
$50.0$ & --    & --    & \bestcell{249} & --          \\
\bottomrule
\end{tabular}
}%
\vspace{-10pt}
\end{wraptable}

The validation checkpoints show a compute advantage before the final model selection point. \Cref{tab:convergence} reports the first validation checkpoint where each construction crosses a fixed rubric-reward threshold on Qwen3-VL-4B/MM. For readability, we report this analysis on a single illustrative setting (Qwen3-VL-4B on MM); the same ordering holds on the other base policies and on HB, with comparable speed-ups, and we treat the per-setting numbers reported in \cref{tab:mm_main_results,tab:hb_results} as the canonical efficiency reference. POW3R reaches $46.0$ dev reward at step~$83$, while the static scalar needs $249$ steps and the category-balanced reward needs $332$; it is also the only method to cross a $50.0$ dev threshold within the schedule. The speed-up does not come from a higher learning rate or a different optimizer schedule: every method here shares the same GRPO recipe, the same prompt budget, and the same evaluation checkpointing.
\WFclear

\subsection{Learning dynamics at matched compute}
\label{sec:trajectories}

\begin{figure}[t]
\centering
\includegraphics[width=\linewidth]{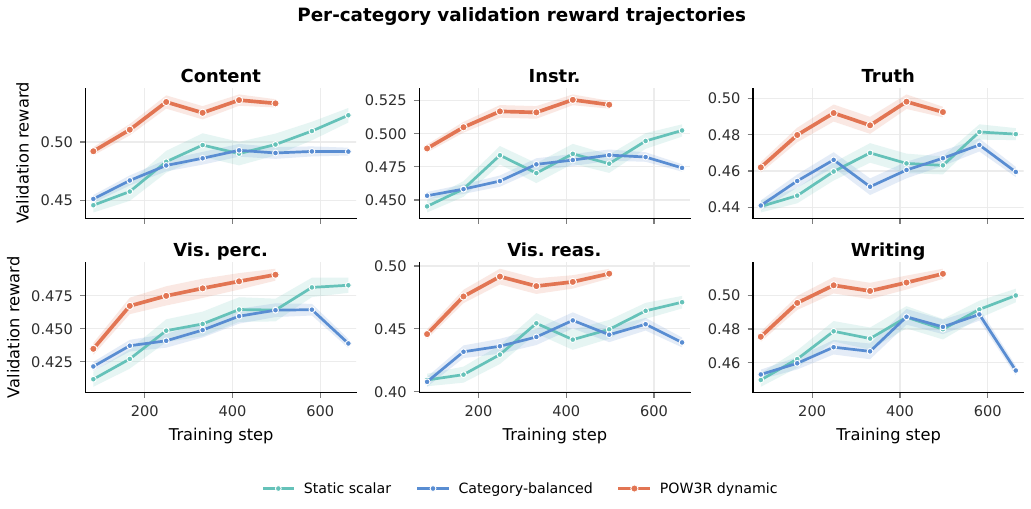}
\vspace{-10pt}
\caption{\textbf{Per-category validation reward trajectories (Qwen3-VL-4B, MM).} POW3R leads on every rubric category throughout training, with the gap opening at the first checkpoint. The largest absolute gains appear on Content, Instruction Following, Truthfulness, Visual Perception, and Visual Reasoning; on Writing Style POW3R still stays above the static baselines, but the three curves stay within roughly a point of each other because most Writing Style criteria are already passed by the base policy and offer little contrastive signal for POW3R to act on.}
\vspace{-12pt}
\label{fig:learning_dynamics}
\end{figure}

\begin{wrapfigure}[14]{r}{0.49\textwidth}
\vspace{-34pt}
\centering
\includegraphics[width=\linewidth]{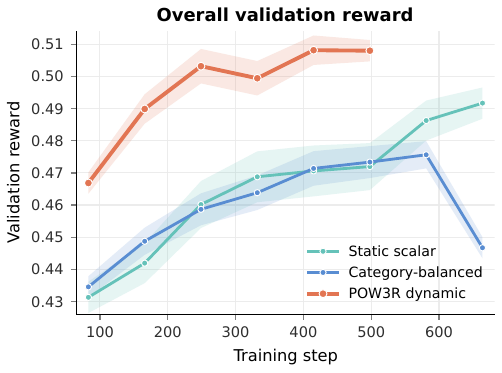}
\vspace{-15pt}
\caption{\textbf{Overall validation reward trajectory (Qwen3-VL-4B, MM).} POW3R separates from the baselines by the first logged checkpoint and stays above through the end of training.}
\label{fig:training_trajectory}
\end{wrapfigure}

The matched-compute curves report validation rubric reward at logged checkpoints, not the per-rollout reward used for training updates. In \cref{fig:training_trajectory}, POW3R separates from both static aggregations by the first logged checkpoint and remains higher through step~$501$; the binary reward never moves above base and is omitted.

The category breakdown in \cref{fig:learning_dynamics} shows that this is not a one-category over-optimization. POW3R improves Content, Instruction Following, Truthfulness, Visual Perception, and Visual Reasoning. Meanwhile the Writing Style reward moves the least because most style criteria are already passed by the base policy and therefore have low rollout variance for POW3R to act on.
\WFclear

\subsection{Where informative updates concentrate}
\label{sec:informative_updates}

\begin{wrapfigure}[18]{r}{0.49\textwidth}
\vspace{-25pt}
\centering
\includegraphics[width=\linewidth]{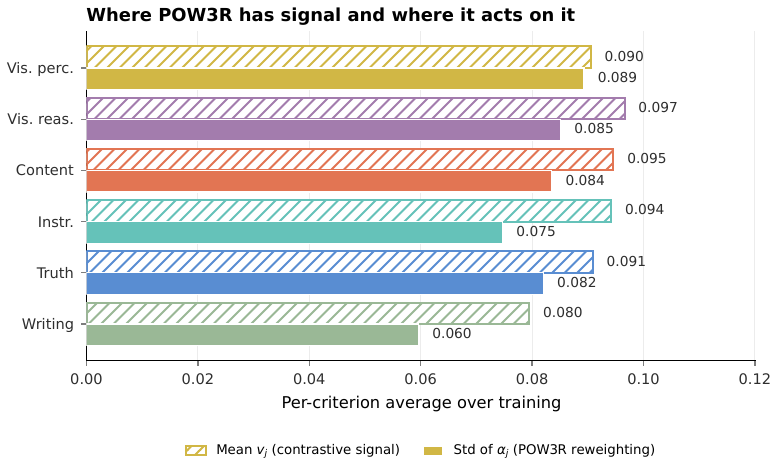}
\vspace{-8pt}
\caption{\textbf{Where POW3R has signal and where it acts on it.} Per-category averages over MM-train criteria, computed over the full training schedule (Qwen3-VL-4B POW3R-dynamic). Hatched: mean rollout variance $v_j$. Solid: within-category standard deviation of POW3R's factors $\alpha_j$. The two co-vary: categories with more contrast get reweighted more.}
\label{fig:informative_updates}
\vspace{-10pt}
\end{wrapfigure}

\Cref{fig:informative_updates} reports two per-category averages from the POW3R-dynamic run, taken over the full training schedule: the mean rollout variance $v_j$ across criteria in the category (the contrastive signal POW3R has to work with) and the within-category standard deviation of POW3R's factors $\alpha_j$ (how much POW3R differentiates criteria inside that category). The two co-vary in the expected direction: Writing Style sits lowest on both ($v_j{\approx}0.080$, $\mathrm{std}(\alpha_j){\approx}0.060$), the visual and reasoning categories sit at the top of $v_j$, and the ordering of $\mathrm{std}(\alpha_j)$ tracks $v_j$ to within rounding. POW3R's category mass stays uniform by construction (\cref{eq:cat_reward}); the chart shows the \emph{within-category} reweighting that actually transmits POW3R's mechanism into the GRPO advantage, and confirms it scales with the rollout-level contrast each category exposes.
\WFclear

\vspace{-10pt}

\section{Limitations}
\label{sec:limitations}

\paragraph{Reliance on LLM judges.}
POW3R reads every rubric criterion through an LLM judge: GPT-5.4-nano for training and GPT-5.4-mini for held-out evaluation, calibrated against a high-effort reference judge with $91$--$94\%$ verdict agreement (\cref{app:judges}). Because both training rewards and reported scores flow through these judges, exact numbers can shift with judge, prompt template or sampling choice and POW3R's policy-aware factors are themselves estimated from judge verdicts, so any consistent verdict bias propagates into where training pressure is redirected. The qualitative ordering POW3R $>$ category-balanced $>$ static scalar $>$ binary is, however, stable across our calibration and across two judge tiers.

\vspace{-15pt}

\paragraph{Scarcity of rubric datasets with static human weights.}
Public rubric-RL datasets that combine prompt-specific criteria, integer human weights, multiple high-level categories, and held-out scale are rare. We therefore authored MM in-house and used HealthBench for the text-only setting. These choices leave POW3R untested on domains such as long-form coding feedback, scientific writing, or multilingual instruction following, which can have very different category structures and saturation patterns. Broader open availability of rubric datasets with static weights would let the community measure the limits of policy-aware aggregation more directly.

\vspace{-10pt}

\section{Conclusion and future work}
\label{sec:conclusion}

POW3R separates two roles that static rubric rewards conflate: what should matter in the final answer and what can currently teach the policy. Its policy-aware reward redirects training pressure using rollout-level contrast without changing the rubric target, yielding higher rubric reward, higher strict completion, cleaner two-objective dominance, and faster training across MM, HealthBench, and two model families. The broader takeaway is that rubric aggregation deserves to be treated as a first-class training-time design choice; careful re-weighting of an existing rubric can extract substantially more signal from the same data, judge, and optimizer. We plan to test POW3R with stronger and more diverse judges, longer training schedules, and adaptive rubric generation that co-evolves with the policy.

\paragraph{Use of generative AI.}
A large language model was used during manuscript preparation for editing, structural revision, and refactoring of figure-generation code. All technical content -- experimental design, results, equations, and citations -- was produced and verified by the authors.

\small
\bibliographystyle{unsrtnat}
\bibliography{refs}

\normalsize
\clearpage
\appendix

\section*{Appendix}
\addcontentsline{toc}{section}{Appendix}
\paragraph{Appendix index.}
\begin{itemize}[leftmargin=*, itemsep=1pt, topsep=2pt]
\item \textbf{\Cref{app:datasets}: Dataset and contributor details} -- rubric annotations, contributor demographics, and split details for MM.
\item \textbf{\Cref{app:judges}: Judge selection} -- reference-judge calibration, cost--quality tradeoffs, and operating points.
\item \textbf{\Cref{app:prompts}: System prompts} -- rubric-judge prompts used for training and evaluation.
\item \textbf{\Cref{app:hyperparams}: Training hyperparameters} -- consolidated GRPO and POW3R settings shared across all runs.
\item \textbf{\Cref{app:rubric_categories}: Rubric categories with example criteria} -- the six high-level MM rubric categories with two representative criteria each, drawn verbatim from MM-train.
\item \textbf{\Cref{app:qualitative}: Qualitative examples} -- four MM held-out tasks with side-by-side outputs from the base policy, the static baselines, and POW3R.
\end{itemize}

\section{Dataset and contributor details}
\label{app:datasets}

This section provides additional details regarding the data collection pipeline, contributor demographics, and dataset splits for the multimodal dataset.

\subsection{Rubric annotations and weights}
\label{app:rubric_annotations}
Each MM example contains prompt-specific rubric criteria annotated with a category, an explicit/implicit tag, an objective/subjective tag, and a static human weight. \textbf{Explicit} criteria are directly stated by the user prompt or image-grounded instruction, such as a required object, format, count, comparison, or named constraint. \textbf{Implicit} criteria are not quoted verbatim from the prompt but are required for a high-quality answer, such as avoiding unsupported claims, explaining uncertainty, or preserving the intended audience and tone. \textbf{Objective} criteria can be judged by a concrete fact, visual evidence, format constraint, or measurable condition; \textbf{subjective} criteria require a quality judgment about clarity, helpfulness, style, or appropriateness. Weights are integer importance ratings in $\{1,\ldots,5\}$ assigned during rubric authoring: low weights mark helpful but nonessential details, middle weights mark important response-quality criteria, and high weights mark criteria whose failure substantially changes answer correctness or usefulness. HealthBench retains its native point values and categories; the conversion used during training and the unchanged evaluation recipe are described in \cref{app:hb_signed}.

\subsection{Contributor demographics and selection}
MM was authored by a vetted pool of English-proficient contributors selected for experience with analytical writing, visual reasoning, and rubric-based evaluation. We omit fine-grained vendor-identifying demographics, collection dates, geographic distribution, and job-title breakdowns from this release; aggregate quality-control statistics, redacted instructions, and the contributor screening criteria are available on request.

\subsection{Dataset funnel and splits}
Our QA and review process served as a strict filtering funnel. Examples that failed independent review or did not meet dataset-level quality criteria were discarded before the final $10{,}000$ examples were split into training, development, and test sets.

\subsection{Signed HealthBench criteria during training and evaluation}
\label{app:hb_signed}
HealthBench criteria carry signed integer point values: a positive $w_j$ rewards a desired behavior, a negative $w_j$ penalizes an undesired behavior (e.g.\ unsafe advice). POW3R's category-normalized aggregation (Eqs.~(\ref{eq:cat_reward})--(\ref{eq:rc_reward})) assumes non-negative weights, since within-category normalizers divide by $\sum_{j} w_j$ and the dynamic-factor renormalization treats $w_j$ as a non-negative prior. Motivated additionally by \citet{rubrichub}'s finding that positive-only rubric training outperforms inclusion of negative penalties, we convert signed HB criteria into an equivalent good-behavior representation for training. To use POW3R on HB without changing the optimizer or the rubric set, we apply a one-step \emph{good-behavior conversion} \emph{only at training time}: for every criterion with $w_j < 0$ we substitute the equivalent non-negative criterion that flips the judge verdict and uses $|w_j|$ as the weight, so $s'_j = 1 - s_j$ and $w'_j = |w_j|$. After this rewrite, $s'_j = 1$ means ``the response avoids the prohibited behavior,'' all weights are positive integers, and POW3R's category normalization and dynamic factors are well-defined. The judge prompt and category labels for those criteria are adjusted accordingly so the LLM judge scores ``avoidance'' rather than ``occurrence.'' Evaluation never sees this rewrite: the HB \emph{Strict} column in \cref{tab:hb_results}, the HB overall score, and every other reported HB number are computed under HealthBench's original scoring script with its native signed point values and unchanged criterion text. The good-behavior conversion only affects what POW3R's category normalizer and dynamic factors see during training, so reported HB results are directly comparable to HealthBench's own protocol.

\section{Judge selection}
\label{app:judges}

The rubric judge is a fixed component of the reward computation, so its choice has a direct effect on every downstream comparison in this paper. We therefore treat judge selection as a first-class experimental decision: we calibrate a small number of candidate judges against a high-effort reference judge, then choose two operating points -- one for high-throughput training, one for slower but more careful held-out evaluation -- using a single explicit cost--quality trade-off.

\paragraph{Reference judge.}
We use GPT-5.4 with high-effort reasoning and the per-criterion explanation kept on as our reference (``gold'') judge~\citep{openai_gpt54}. This is the strongest model available to us through our LLM provider; high-effort reasoning gives it the most slack for careful per-criterion judgments, and keeping the explanation on prevents it from short-circuiting to a verdict. We acknowledge that a single reference judge cannot establish ground truth in an absolute sense; we therefore treat the agreement numbers below as a calibration of cheaper judges \emph{against} the strongest judge we can afford to run, not as ground-truth accuracy.

\paragraph{Calibration set.}
We sample $1{,}000$ stratified task--rollout--rubric triples from MM-train and re-judge each one with the reference judge and every candidate judge configuration, varying model, reasoning effort, and explanation mode. Sampling is stratified across the six rubric categories and the rubric weight bins $w \in \{1\!-\!3, 4\!-\!5\}$, so the calibration covers easy- and hard-to-judge regions evenly. All candidates score the same $1{,}000$ items, so per-judge agreement is paired and apples-to-apples.

\paragraph{Per-rubric vs. per-category judging.}
We also tested issuing one combined judge call per rollout category instead of per rollout criterion. The per-category mode is roughly $C$-times cheaper but rank-correlation against the reference drops appreciably and the judge tends to merge or skip individual rubrics whose verdicts ought to remain independent. We therefore keep the per-rubric formulation throughout the paper.

\paragraph{Per-rubric judge calibration.}
\Cref{tab:judges} reports pairwise verdict agreement for each candidate judge configuration. The \emph{reference} row shows the reference judge's own agreement against a small human-rated subset of the calibration items (used to set a quality ceiling at $95.4\%$); every other row reports agreement of the candidate judge against the reference judge on the full $1{,}000$-item calibration set. Mini-class judges (GPT-5.4-mini) at medium effort with explanation track the reference within ${\sim}2$ points at roughly $1/8$ of the reference cost; nano-class judges (GPT-5.4-nano) at medium effort with explanation lose another ${\sim}2$ points but are an additional ${\sim}10\times$ cheaper, with rank fidelity preserved when the explanation is kept on.

\begin{table}[ht]
\centering
\caption{Calibration of candidate per-rubric judges against the reference judge (GPT-5.4 with high-effort reasoning and explanation). The reference row's $95.4\%$ entry is the reference judge's agreement with a human-rated subset; every other row is agreement of that candidate judge against the reference on the same $1{,}000$ stratified items from MM-train. Bold rows mark the operating points used in the paper.}
\label{tab:judges}
\small
\setlength{\tabcolsep}{6pt}
\begin{tabular}{llccr}
\toprule
Judge model         & Reasoning  & Explanation & Agreement (\%)         & Cost / 1k (USD) \\
\midrule
GPT-5.4 (reference) & high       & yes & $95.4$                 & $12.50$ \\
\midrule
GPT-5.4-mini        & high       & yes & $94.1$                 & $\phantom{0}2.19$ \\
GPT-5.4-mini        & high       & no  & $93.7$                 & $\phantom{0}1.79$ \\
\textbf{GPT-5.4-mini}        & \textbf{medium}     & \textbf{yes} & $\mathbf{93.6}$        & $\phantom{0}1.52$ \\
GPT-5.4-mini        & medium     & no  & $93.4$                 & $\phantom{0}1.14$ \\
GPT-5.4-mini        & low        & yes & $89.5$                 & $\phantom{0}1.06$ \\
GPT-5.4-mini        & low        & no  & $87.9$                 & $\phantom{0}0.76$ \\
\midrule
GPT-5.4-nano        & high       & yes & $91.8$                 & $\phantom{0}0.13$ \\
\textbf{GPT-5.4-nano}        & \textbf{medium}     & \textbf{yes} & $\mathbf{91.4}$        & $\phantom{0}0.12$ \\
GPT-5.4-nano        & medium     & no  & $91.4$ (rank break)    & $\phantom{0}0.08$ \\
GPT-5.4-nano        & low        & yes & $88.0$                 & $\phantom{0}0.10$ \\
GPT-5.4-nano        & low        & no  & $82.6$                 & $\phantom{0}0.07$ \\
\bottomrule
\end{tabular}
\end{table}

Two notes from the table. First, removing the per-criterion explanation at the nano scale gives the same head-line agreement number but breaks rank correlation between the candidate's per-model rubric pass rate and the reference's: agreements at the right level on average, but disagreements that flip per-model rankings. We therefore always keep the explanation on. Second, mini-class judges are roughly $12\times$ more expensive than nano-class judges per $1{,}000$ calls, but agreement gains from mini-medium-with-explanation over nano-medium-with-explanation are ${\sim}2$ percentage points -- non-trivial for held-out evaluation, but not worth $12\times$ the cost on every training rollout.

\paragraph{Operating points used in the paper.}
Based on the table above, we use \textbf{GPT-5.4-nano + medium reasoning + explanation} ($91.4\%$ agreement, \$0.12 / $1{,}000$ judgments) for training rewards, and \textbf{GPT-5.4-mini + medium reasoning + explanation} ($93.6\%$ agreement, \$1.52 / $1{,}000$ judgments) for held-out evaluation rewards. Using the more accurate model only at evaluation time also reduces judge--training entanglement: training never sees the verdict distribution it will be evaluated against.

\section{System prompts}
\label{app:prompts}

This appendix collects the system prompts used by the rubric judges referenced in \cref{sec:setup}. Both judges share the same single-criterion template; only the underlying model differs (GPT-5.4-nano with medium reasoning at training time, GPT-5.4-mini with medium reasoning at evaluation time). Both are queried at temperature $T{=}1.0$ with up to $2048$ completion tokens.

\paragraph{Per-criterion rubric judge.}
Given a sampled response and a single rubric criterion (with its category, explicit/implicit and objective/subjective annotations, and human weight), the judge returns a JSON object with a one-sentence reasoning and a binary verdict.

\begin{promptbox}
\footnotesize\ttfamily\raggedright
Does the response satisfy this rubric criterion? Evaluate ONLY this criterion.\\[2pt]
Rules:\\
- "Explicit" criteria must be directly addressed. "Implicit" may be inferred.\\
- "Objective" = factual pass/fail. "Subjective" = quality judgment.\\
- Weight is context only --- does not affect your pass/fail decision.\\
- Minor phrasing/formatting differences are OK if substance is correct.\\
- OCR/text recognition criteria require exact text --- "EXIST" $\neq$ "EXIT".\\[2pt]
Examples:\\
- Rubric: "Identify chair material as wood" / Response: "a wooden chair" $\rightarrow$ \{"reasoning": "Explicitly identifies wood.", "criteria\_met": true\}\\
- Rubric: "Graph shows decreasing trend after 2020" / Response: "steady growth" $\rightarrow$ \{"reasoning": "Claims growth, not decrease.", "criteria\_met": false\}\\
- Rubric: "List $\geq$3 differences" / Response: "one difference" $\rightarrow$ \{"reasoning": "Only 1 of 3 required.", "criteria\_met": false\}\\
- Rubric: "Read sign as `EMERGENCY EXIT"' / Response: "\,`EMERGENCY EXIST'\," $\rightarrow$ \{"reasoning": "EXIST $\neq$ EXIT, OCR must be exact.", "criteria\_met": false\}\\[2pt]
Rubric:\\
- Title: \{rubric\_title\}\\
- Category: \{rubric\_category\}\\
- \{explicit\_implicit\} | \{objective\_subjective\} | Weight: \{rubric\_weight\}\\
- Criteria: \{rubric\_rationale\}\\[2pt]
Response:\\
\{response\}\\[2pt]
Return ONLY valid JSON. "reasoning" BEFORE "criteria\_met".\\
\{"reasoning": "$\langle$one sentence$\rangle$", "criteria\_met": true/false\}
\end{promptbox}

The criterion-level judge prompt above is the prompt template used by the training (GPT-5.4-nano) and held-out evaluation (GPT-5.4-mini) judges, and is what is needed to reproduce the reward calculations reported in the main paper.

\paragraph{Verdict-only per-criterion variant (calibration only).}
This prompt is identical in intent to the per-criterion judge above, but asks the judge to emit only the binary verdict, without the one-sentence reasoning field. \Cref{app:judges} reports that this variant matches the with-explanation variant on aggregate verdict agreement but loses per-model rank fidelity, which is why we keep the explanation on in the operating points used in the paper.

\begin{promptbox}
\footnotesize\ttfamily\raggedright
Does the response satisfy this criterion? Evaluate ONLY this criterion.\\
"Explicit" = directly addressed. "Implicit" = may be inferred.\\
OCR/text recognition = exact match required.\\[2pt]
Rubric:\\
- Title: \{rubric\_title\} | Category: \{rubric\_category\}\\
- \{explicit\_implicit\} | \{objective\_subjective\} | Weight: \{rubric\_weight\}\\
- Criteria: \{rubric\_rationale\}\\[2pt]
Response:\\
\{response\}\\[2pt]
Return ONLY valid JSON: \{"criteria\_met": true\} or \{"criteria\_met": false\}
\end{promptbox}

\paragraph{Per-category batched variant (calibration only).}
This prompt scores all rubric criteria for a single response in one call by asking the judge to evaluate each criterion independently and emit a single weight-normalized $[0,1]$ score. We use it only in the cost--quality calibration in \cref{app:judges}; in the paper itself every rubric criterion is judged one at a time.

\begin{promptbox}
\footnotesize\ttfamily\raggedright
Score the response against all rubrics below. Evaluate each independently as pass/fail.\\[2pt]
Rules:\\
- "Explicit" = directly addressed. "Implicit" = may be inferred.\\
- "Objective" = factual. "Subjective" = quality judgment.\\
- OCR/text recognition criteria require exact text match.\\
- score = sum(weight of PASSED rubrics) / \{total\_weight\}, between 0.0 and 1.0.\\[2pt]
Rubrics:\\
\{rubrics\_text\}\\[2pt]
Total weight: \{total\_weight\}\\[2pt]
Response:\\
\{response\}\\[2pt]
Return ONLY valid JSON. "reasoning" BEFORE "score".\\
\{"reasoning": "$\langle$which rubrics passed/failed$\rangle$", "score": $\langle$0.0--1.0$\rangle$\}
\end{promptbox}

All three prompts are taken verbatim from our training and evaluation code; only formatting (line breaks, font, and the surrounding box) is adjusted for the paper.

\section{Training hyperparameters}
\label{app:hyperparams}

\Cref{tab:hyperparams} consolidates the GRPO and POW3R hyperparameters used for every training run reported in this paper. The same values are used across all base policies (Qwen3-VL-4B, Qwen3-VL-8B, Gemma3-4B for MM; Qwen3-4B, Qwen3-8B, Gemma3-4B for HB) and across the Binary, Static scalar, Category-balanced, and POW3R dynamic reward constructions.

\begin{table}[h]
\centering
\caption{Training hyperparameters shared across all runs in \cref{tab:mm_main_results,tab:hb_results}.}
\label{tab:hyperparams}
\small
\setlength{\tabcolsep}{8pt}
\begin{tabular}{lll}
\toprule
\textbf{Group} & \textbf{Hyperparameter} & \textbf{Value} \\
\midrule
\multirow{6}{*}{GRPO objective}
 & Rollouts per prompt-group ($G$)            & $16$ \\
 & Sampling temperature ($T$)                 & $1.0$ \\
 & Maximum completion length                  & $3584$ tokens \\
 & KL coefficient ($\beta$)                   & $0.1$ \\
 & PPO clip range ($\varepsilon$)             & $0.2$ \\
 & Gradient clip ($\mathrm{max\_grad\_norm}$) & $0.5$ \\
\midrule
\multirow{4}{*}{Optimizer}
 & Optimizer                                  & AdamW \\
 & Learning rate                              & $3 \!\times\! 10^{-7}$ \\
 & LR schedule                                & constant \\
 & Weight decay                               & $0.0$ \\
\midrule
\multirow{5}{*}{Batch \& schedule}
 & Per-device train batch size                & $1$ \\
 & Gradient accumulation steps                & $4$ \\
 & Maximum GRPO steps                         & $664$ \\
 & Validation interval                        & every $83$ steps \\
 & Checkpoint selection                       & val-best per method \\
\midrule
\multirow{6}{*}{POW3R ($R_{\text{dyn}}$)}
 & Variance floor ($\alpha_{\min}$)           & $0.67$ \\
 & Variance ceiling ($\alpha_{\max}$)         & $1.5$ \\
 & Numerical stability ($\epsilon$)           & $10^{-4}$ \\
 & Smoothing weight ($\lambda$)               & $0.5$ \\
 & EMA coefficient ($\beta_{\mathrm{ema}}$)   & $0.2$ \\
 & Minimum valid rollout fraction             & $0.75$ \\
\midrule
\multirow{4}{*}{Hardware \& precision}
 & Compute                                    & $1{\times}$ node, $8{\times}$ H100 GPUs \\
 & Distributed strategy                       & DeepSpeed ZeRO-3 \\
 & Numerical precision                        & BF16 \\
 & Activation checkpointing                   & enabled \\
\midrule
\multirow{3}{*}{Judges}
 & Training judge                             & GPT-5.4-nano, medium reasoning, explanation on \\
 & Evaluation judge                           & GPT-5.4-mini, medium reasoning, explanation on \\
 & Judge sampling                             & $T{=}1.0$, max $2048$ completion tokens \\
\bottomrule
\end{tabular}
\end{table}

\section{Rubric categories with example criteria}
\label{app:rubric_categories}

MM rubrics are tagged with one of six high-level quality categories: \emph{Visual perception}, \emph{Visual reasoning}, \emph{Content completeness}, \emph{Instruction following}, \emph{Truthfulness}, and \emph{Writing style / presentation}. Each category captures a different aspect of response quality so that POW3R's category-normalized aggregation (\cref{eq:rc_reward}) can balance grounding, reasoning, completeness, instruction-following, factual correctness, and presentation criteria within the same prompt. Below we list each category, a one-line operational definition, and two representative criteria taken verbatim from MM-train ($7$--$10$ such criteria typically populate one task across these six categories; see \cref{tab:dataset}). The bracketed tag on each criterion is \texttt{[weight | explicit/implicit | objective/subjective]} per the schema in \cref{app:rubric_annotations}.

\begin{catbox}{scaleGold}
\textbf{Visual perception.} What the image directly shows: objects, text, named entities, counts, and other content that can be read off the image without further inference.
\begin{itemize}[leftmargin=2em, itemsep=2pt, topsep=2pt]
\item \texttt{[5 | Explicit | Objective]} ``The response must identify at least two of the visible players on the court by specifying their names (e.g., LeBron James, Anthony Davis) and their respective teams (e.g., Lakers or Rockets).''
\item \texttt{[4 | Explicit | Objective]} ``The response must identify the stores from the letter by listing the following stores: K-Mart, Toys `R' Us, Sears, Roebuck \& Co., and Montgomery Ward.''
\end{itemize}
\end{catbox}

\begin{catbox}{scalePurple}
\textbf{Visual reasoning.} Inferences that require combining visual cues, such as version identification from interface details, structural interpretation of diagrams and circuits, or reasoning about visual context.
\begin{itemize}[leftmargin=2em, itemsep=2pt, topsep=2pt]
\item \texttt{[4 | Explicit | Objective]} ``The response must identify the specific version of NBA 2K shown in the screenshot (2K20) based on visual elements such as graphics, court layout, or interface design unique to that version.''
\item \texttt{[4 | Explicit | Objective]} ``The response must infer who wrote the initial request and what their request was by referencing the apology in the P.S.\ to Laura about LEGO being unable to send product brochures.''
\end{itemize}
\end{catbox}

\begin{catbox}{scaleTerracotta}
\textbf{Content completeness.} Whether the response covers every element the prompt asks for, with all required items, sub-items, comparisons, or sections present.
\begin{itemize}[leftmargin=2em, itemsep=2pt, topsep=2pt]
\item \texttt{[5 | Explicit | Objective]} ``The response must identify all of the labeled objects in the image: Algenib, Deneb Kaitos, Fische, Fomalhaut, Jupiter, Merkur, Neptun, Sonne, and Wasserman.''
\item \texttt{[4 | Explicit | Objective]} ``The response must explain how each proposed power factor correction method affects both reactive and active power by discussing the change in the circuit's power characteristics.''
\end{itemize}
\end{catbox}

\begin{catbox}{scaleTeal}
\textbf{Instruction following.} Whether the response respects explicit constraints in the prompt: ordering, sectioning, length, format, and procedural requirements.
\begin{itemize}[leftmargin=2em, itemsep=2pt, topsep=2pt]
\item \texttt{[2 | Explicit | Objective]} ``The response must identify the labeled entities in this order: Algenib, Deneb Kaitos, Fische, Fomalhaut, Jupiter, Merkur, Neptun, Sonne, and Wasserman.''
\item \texttt{[3 | Explicit | Objective]} ``The response must provide the answers in 2 sections.''
\end{itemize}
\end{catbox}

\begin{catbox}{scaleBlue}
\textbf{Truthfulness.} Whether claims in the response are accurate -- correct identification of named entities, correct relationships, correct intermediate computations, and absence of unsupported claims.
\begin{itemize}[leftmargin=2em, itemsep=2pt, topsep=2pt]
\item \texttt{[5 | Explicit | Objective]} ``The response must state the television series shown in the image by identifying the show as Avatar: The Last Airbender.''
\item \texttt{[5 | Explicit | Objective]} ``The response must analyse the impedance change by explaining how inductance affects impedance compared to resistance.''
\end{itemize}
\end{catbox}

\begin{catbox}{scaleSage}
\textbf{Writing style / presentation.} Formatting, organization, register, and adherence to specific presentation requirements such as bullet lists, bolding, or audience-appropriate language.
\begin{itemize}[leftmargin=2em, itemsep=2pt, topsep=2pt]
\item \texttt{[4 | Explicit | Objective]} ``The response must order these alphabetically by reading their names and organising them into an alphabetised bullet list.''
\item \texttt{[2 | Explicit | Subjective]} ``The response should provide the answers in a high school student's language level by using clear and simple terms and sentences.''
\end{itemize}
\end{catbox}

\section{Qualitative examples}
\label{app:qualitative}

We present four MM held-out tasks from the Qwen3-VL-4B test set, drawn from four different task families (Spatial Reasoning, Visual Perception, Visual Reasoning on a polar curve, and a Visual-grounded math worksheet). For each task we show the image, the prompt, what the rubric expects, and the outputs of the base policy together with the three trained settings ($R_{\text{scalar}}$, $R_{\text{cat}}$, $R_{\text{dyn}}$). Pass counts report the number of rubric criteria satisfied out of the task's total criterion count, while $\textsf{strict}{=}\textsf{True}$ indicates that every criterion flagged as \emph{required} (the subset that defines strict completion in \cref{tab:mm_main_results,tab:hb_results}) was satisfied; a method can therefore be $\textsf{strict}{=}\textsf{True}$ while leaving optional criteria unsatisfied. Bold marks the substantive error or correct decision that distinguishes the methods. Reading these side by side highlights the kinds of grounding, completeness, and naming behaviors that POW3R recovers relative to the static baselines.

\needspace{6cm}
\begin{promptbox}
\textbf{Example 1: Spatial reasoning on a Tokyo metro map.}
\\[3pt]
\begin{minipage}[t]{0.36\textwidth}
\vspace{0pt}
\centering
\includegraphics[width=\linewidth]{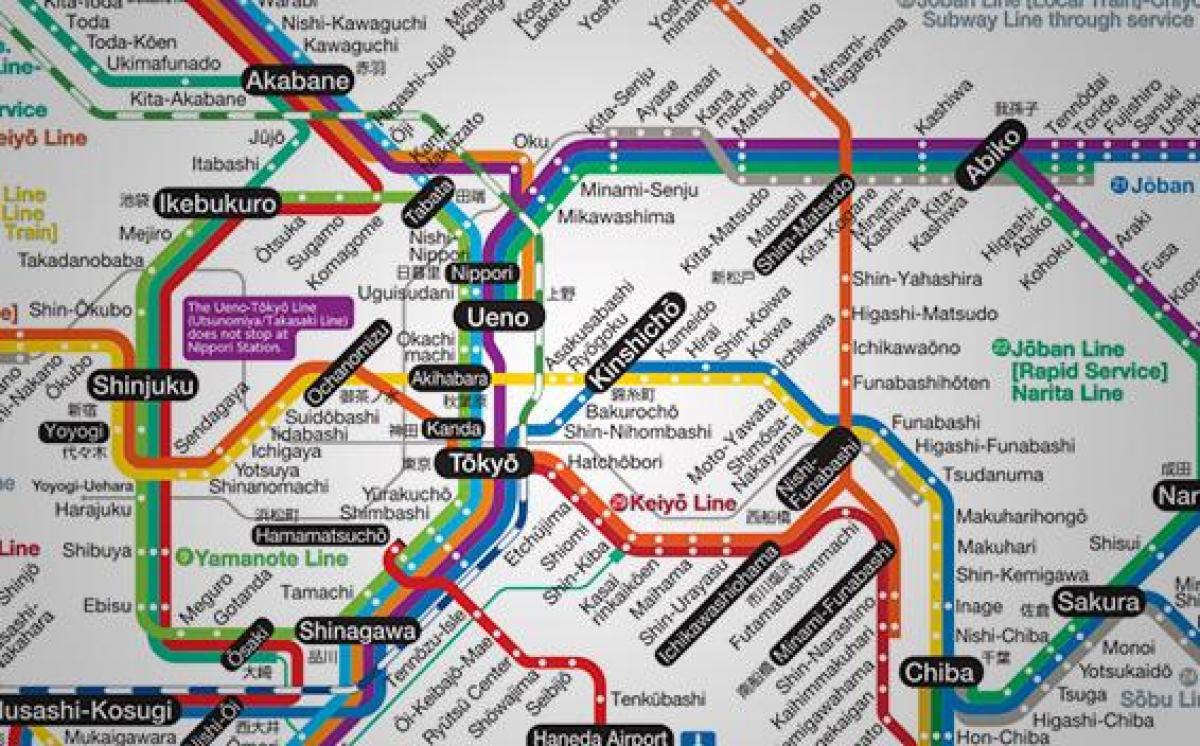}
\end{minipage}
\hfill
\begin{minipage}[t]{0.60\textwidth}
\vspace{0pt}\small
\textbf{Prompt:} \emph{Based on the layout and connectivity of the transit system shown in the image, what station is the most central hub for transfers, and what factors contribute to its importance?}

\smallskip\textbf{Required answer (rubric):} Tokyo Station -- visually centered on the map and the highest-degree transfer node.

\smallskip\textbf{Base / $R_{\text{scalar}}$ / $R_{\text{cat}}$:} all confidently answer ``\textbf{Ikebukuro Station}'', a peripheral upper-left node, and hallucinate line intersections that are not in the image. \textit{(0/3 required rubrics passed for each method.)}

\smallskip\textbf{POW3R ($R_{\text{dyn}}$):} ``Tokyo Station, located at the center of the map. Multiple major lines converge here.'' \textit{(3/3 passed.)}
\end{minipage}
\end{promptbox}

\needspace{6cm}
\begin{promptbox}
\textbf{Example 2: Visual perception on a meme image (animal identification).}
\\[3pt]
\begin{minipage}[t]{0.30\textwidth}
\vspace{0pt}
\centering
\includegraphics[width=\linewidth]{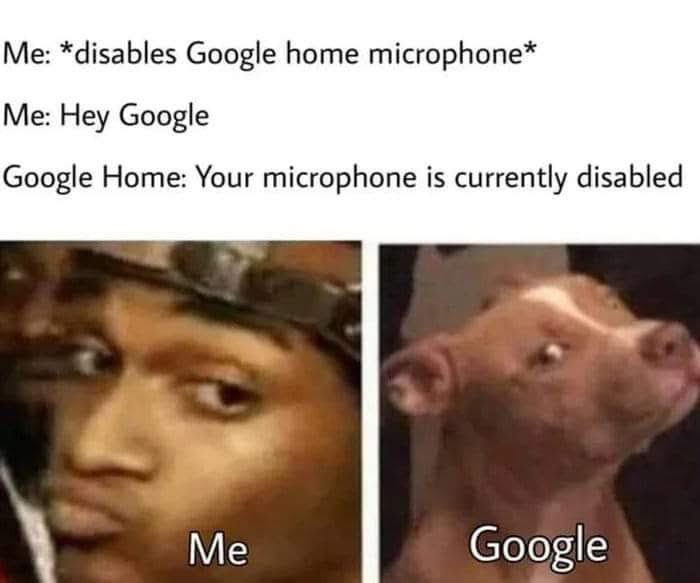}
\end{minipage}
\hfill
\begin{minipage}[t]{0.66\textwidth}
\vspace{0pt}\small
\textbf{Prompt:} \emph{Can you explain why the google home is represented by a dog? What does that have to do with the context?}

\smallskip\textbf{Required answer (rubric):} Acknowledge the dog, describe its expression, and connect the dog's indifferent look to a Google-Home obedience-but-misunderstanding joke.

\smallskip\textbf{Base / $R_{\text{scalar}}$ / $R_{\text{cat}}$:} all argue with the user that the image is \textbf{a pig, not a dog}, and proceed to explain a different ``pig'' meme. The visual misperception cascades into the rest of the answer. \textit{(0/5 required rubrics passed.)}

\smallskip\textbf{POW3R ($R_{\text{dyn}}$):} accepts the prompt's framing, identifies the dog and its indifferent expression, and ties the dog's obedience-but-not-understanding pose to the Google Home punchline. \textit{(5/5 passed.)}
\end{minipage}
\end{promptbox}

\needspace{6cm}
\begin{promptbox}
\textbf{Example 3: Visual reasoning on a polar-curve identification.}
\\[3pt]
\begin{minipage}[c]{0.30\textwidth}
\centering
\includegraphics[width=\linewidth]{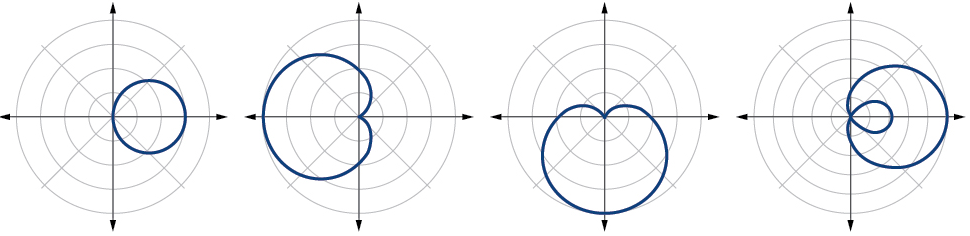}
\end{minipage}
\hfill
\begin{minipage}[c]{0.66\textwidth}
\small
\textbf{Prompt:} \emph{How about the second shape? What's its name?}

\smallskip\textbf{Required answer (rubric):} The second shape is a \textbf{cardioid} ($r = a(1 + \cos\theta)$), recognized from its single rightward cusp and heart shape.

\smallskip\textbf{Base / $R_{\text{scalar}}$ / $R_{\text{cat}}$:} hedge between ``rose curve / lima\c{c}on / heart curve'' or pivot to a generic discussion of polar curves \textbf{without naming the specific shape}. \textit{(0--3/8 rubrics passed; strict $=$ False.)}

\smallskip\textbf{POW3R ($R_{\text{dyn}}$):} ``This is a \textbf{cardioid}, $r = a(1+\cos\theta)$, identifiable by the single cusp on the right and the symmetric heart-shaped contour.'' \textit{(5/8 passed; strict $=$ True.)}
\end{minipage}
\end{promptbox}

\needspace{6cm}
\begin{promptbox}
\textbf{Example 4: Visual-grounded math worksheet generation.}
\\[3pt]
\begin{minipage}[t]{0.32\textwidth}
\vspace{0pt}
\centering
\includegraphics[width=\linewidth]{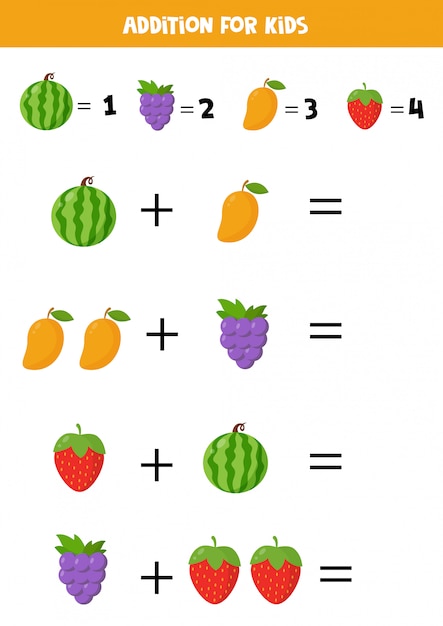}
\end{minipage}
\hfill
\begin{minipage}[t]{0.64\textwidth}
\vspace{0pt}\small
\textbf{Prompt:} \emph{Could you make a new worksheet of 4 sums with these fruit and values that use division, multiplication, and subtraction where each of the answers are 4?}

\smallskip\textbf{Required answer (rubric):} Four arithmetic problems using fruits with assigned values (\,\includegraphics[height=8pt]{figures/examples/67c2439b2082f0ac0b990e2b.jpg}\,$\to 1,2,3,4$); every answer must equal $4$; mix of $\div$, $\times$, $-$.

\smallskip\textbf{Base / $R_{\text{scalar}}$ / $R_{\text{cat}}$:} produce four sums but \textbf{at least one answer is wrong} (e.g.\ ``Mango$-$Watermelon\,$=$\,$2$'' presented as a target-$4$ problem) and the worksheet self-corrects mid-stream rather than committing. \textit{(4--5/13 rubrics passed; strict $=$ False.)}

\smallskip\textbf{POW3R ($R_{\text{dyn}}$):} four clean problems each evaluating to $4$, mixed across $\div$, $\times$, $-$, with the requested fruit-value substitution intact. \textit{(8/13 passed; strict $=$ True.)}
\end{minipage}
\end{promptbox}

\end{document}